\documentclass{bmvc2k}

\usepackage{graphicx}
\usepackage{multirow}
\usepackage{algorithm}
\usepackage{bm}
\usepackage{mathrsfs}
\usepackage{tikz}
\usepackage{comment}
\usepackage{color}
\usepackage{soul}
\usepackage{amssymb}
\usepackage{subcaption}
\usepackage{graphicx}
\usepackage{wrapfig, lipsum}

\newcommand{\method}[1]{\ifthenelse{\equal{#1}{full}}{multiple branch video-text alignment model}{EVA}}

\newcommand{\vparagraph}[1]{\vspace{-0em}\paragraph{#1}}

\usepackage{pifont}%
\newcommand{\cmark}{\ding{51}}%
\newcommand{\xmark}{\ding{55}}%

\title{Hybrid-Learning Video Moment Retrieval across Multi-Domain Labels}

\addauthor{Weitong Cai}{weitong.cai@qmul.ac.uk}{1}
\addauthor{Jiabo Huang}{jiabo.huang@qmul.ac.uk}{1}
\addauthor{Shaogang Gong}{s.gong@qmul.ac.uk}{1}

\addinstitution{
 Computer Vision Group, \\
 School of Electronic Engineering and Computer Science, \\
 Queen Mary University of London, \\ London E1 4NS, UK
}

\runninghead{W. Cai, J. Huang, S. Gong}{Hybrid-Learning Video Moment Retrieval}

\def\eg{\emph{e.g}\bmvaOneDot}

\def\etal{\emph{et al}\bmvaOneDot}

\begin{document}

\maketitle

\begin{abstract}
Video moment retrieval (VMR)
is to search for a visual temporal moment in an untrimmed raw video
by a given text query description (sentence).
Existing studies either start from collecting exhaustive frame-wise annotations
on the temporal boundary of target moments (fully-supervised),
or learn with only the video-level video-text pairing labels (weakly-supervised).
The former is poor in generalisation to unknown concepts and/or novel scenes
due to 
restricted dataset scale and diversity under expensive annotation costs;
the latter is subject to visual-textual mis-correlations from incomplete labels.
In this work,
we introduce a new approach called \textit{hybrid-learning video moment retrieval}
to solve the problem 
by 
knowledge transfer through 
adapting the video-text matching relationships
learned from a fully-supervised source domain
to a weakly-labelled target domain when they do not share 
a common label space.
Our aim is to explore shared universal knowledge
between the two domains in order to improve model learning 
in the weakly-labelled target domain.
Specifically,
we introduce a 
\textit{multiplE branch Video-text Alignment model} (\method{abbr})
that performs
cross-modal (visual-textual) matching information sharing and multi-modal feature alignment
to optimise domain-invariant visual and textual features
as well as per-task discriminative joint video-text representations.
Experiments show 
EVA's effectiveness in exploring temporal segment annotations 
in a source domain to help learn video moment retrieval without temporal labels
in a target domain.
\end{abstract}

\section{Introduction}
\label{sec:intro}

\begin{figure}[t]
   \centering
   \includegraphics[width=\textwidth]{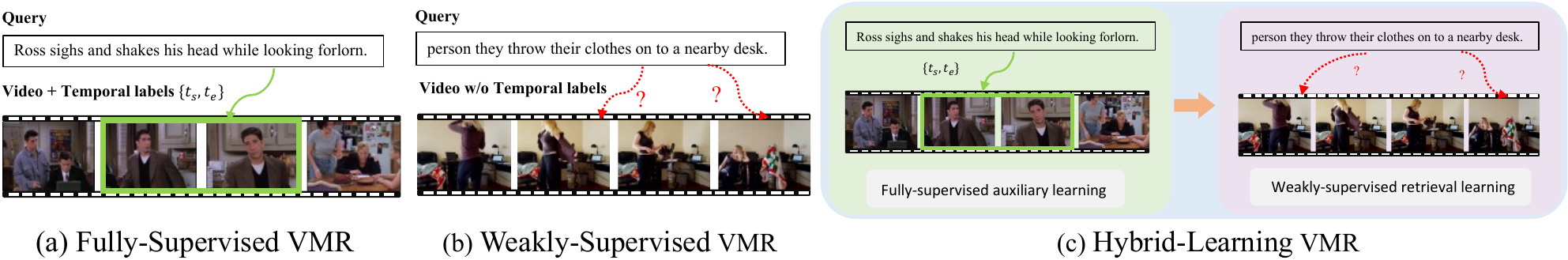}
   \caption{VMR tasks with different supervision settings.
	}
    \label{fig:intro}
\end{figure}

Video moment retrieval (VMR) aims to locate a temporal moment
in a long and untrimmed video
by predicting its start and end time indices
according to a natural language query sentence~\cite{mithun2019tga,duan2018wsdec}.
This problem is intrinsically challenging
as it requires not only 
to derive the 
semantic 
correspondences 
between video and text
but also to
recognise the subtle visual dissimilarity
among different moments in the same unstructured videos
of shared context.
To learn effectively such subtle differences between similar
video segments of different semantic descriptions, the exact temporal
boundaries of target video moments are usually required in model
training. 
However,
such fine-grained 
temporal 
labelling is 
not only 
much more time-consuming but also more ambiguous 
therefore
subjective to 
more noise
than still-image class annotations.

Existing methods take two approaches:
\textbf{(1)}
Fully-supervised
methods~\cite{gao2017tall,anne2017MCN,liu2018attentive,
liu2018cross,yuan2019find,chen2018temporally,jiang2019cross} 
assume the availability of reliable and precise temporal boundary annotations
for model training (Fig.~\ref{fig:intro}(a)).
However,
considering the unaffordable annotation cost,
the publicly available datasets are restricted in both scale and diversity.
This results in poor generalisation of existing fully-supervised methods
to new learning tasks in other domains~\cite{yuan2021ood, bao2022crossscene}.
\textbf{(2)} Alternatively, weakly-supervised
approaches~\cite{ma2020vlanet,lin2020scn,song2020MARN,mithun2019tga,huang2021CRM} 
aims to reduce the annotation cost by model training with only
video-level video-text pairing labels (Fig.~\ref{fig:intro}(b)). 
However, it is harder for weakly-supervised learning to
effectively derive semantically plausible video-text
correspondences
without temporal boundary labelling.

We consider that
semantically plausible video-text correspondences
should be agnostic to domains and can be shared across
learning tasks.
In this work, we formulate a new approach to solving this problem
called \textit{hybrid-learning video moment retrieval},
for performing jointly fully- and weakly-supervised learning
using both a fully-labelled source domain 
and a weakly-labelled target domain simultaneously (Fig.~\ref{fig:intro}(c)).
This hybrid learning approach aims to optimise weakly-supervised
learning of visual-textual correlations in a target domain 
without temporal labels of video moments 
by sharing knowledge on video-text alignment learned from a source domain.
Such a problem is fundamentally challenging due to the 
distribution shift
and vocabulary discrepancy across domains/learning tasks.
Existing datasets are
restricted to specific video scenes
that are distinct between datasets,~\eg, 
ActivityNet-Captions (Anet)~\cite{krishna2017anetcaptions}
involves mostly out-door activities whilst
Charades-STA (Charades)~\cite{gao2017tall} is mostly daily indoor routines.
Videos in different domains 
have different data characteristics
regarding visual and motion patterns, and video durations.
Moreover,
since the natural language queries
are provided without explicit constraint on wording,
existing datasets are different in their vocabulary,
resulting in learning retrieval tasks in different label spaces,~\eg,
TVR~\cite{lei2020tvr} uses
\textit{cat}, \textit{kitty} and \textit{tabby}
to describe the concept of ``cat''
while \textit{tabby} is missing in Anet
and only \textit{cat} is used in Charades.
There are 41\% and 61\% of the words used in TVR and Anet training splits
are shared cross domains.
It is nontrivial to optimise simultaneously visual-textual correlations
between both common vocabularies shared across 
tasks
and similar/related semantics expressed in different words between
  domains.

In this work, we introduce a \textit{multiplE branch
Video-text Alignment model} (\method{abbr}) for hybrid-learning
video moment retrieval 
across tasks
in different domains. 
The key idea is to share the precise video-text temporal label
information in fully-supervised learning in one domain with weakly-supervised learning
in another domain of different data without temporal annotations.
To that end, we formulate two concurrent learning branches with one
weakly-supervised retrieval branch learning from target domain data and another
fully-supervised auxiliary branch learning from a fully-labelled
auxiliary dataset.
For cross-domain cross-modal label sharing in model learning, we apply a
cross-modal attention module to estimate the
correlation between video and query and share this fine-grained matching
knowledge to the weakly-supervised retrieval branch.
Given the distribution shift and label inconsistency
across domains (between datasets), 
we introduce a modality feature alignment constraint to
mitigate feature discrepancies between different data sources in each single
modality. 
This constraint is 
imposed by measuring the maximum mean discrepancy~\cite{xu2019DAVQA,gretton2012MMD}
and applied to both before and after
the cross-modal attention module.
By doing so, the feature spaces of both video-text
modalities are encouraged to share more areas in learning
cross-modal attention to focus on video-text interaction
between domains.
Moreover,
we further deploy a joint-modal domain classifier, combined with
the weakly- and fully-supervised video moment retrieval losses for each retrieval task under different supervision in the two branches.
This is to ensure that the joint
video-text representations are per-task discriminatively optimised.

Our \textbf{contributions} are:
\textbf{(1)}
To our best knowledge,
we make the first attempt at hybrid-learning 
VMR 
to explore jointly
fully-labelled and weakly-labelled domains for
optimising model learning of visual-textual correlations 
in the weakly-labelled domain where temporal labels are not given.
\textbf{(2)}
We introduce a 
multi-branch 
multi-modal formulation to transfer temporal label information as
knowledge in model learning 
across tasks in different domains
by both modal-specific and joint-modal feature alignments.
\textbf{(3)}
Extensive experiments 
show that the video-text alignment knowledge 
derived from the temporal labels 
in a source domain
bring nonnegligible improvements 
to learn activity boundaries in a target domain without manual labels, 
ensuring EVA's competitiveness against the state-of-the-art VMR methods.

\section{Related Works}

\noindent{\bf Fully-supervised video moment retrieval}.
Video moment retrieval was first introduced in~\cite{gao2017tall,anne2017MCN} 
to derive the video-text alignment 
with the help of precise temporal boundary labels 
corresponding to natural language descriptions 
of video activities. 
To that end, MCN~\cite{anne2017MCN} 
generated candidate segments with different lengths 
in various starting points for retrieving.
The studies of this problem mostly focus on either the multi-modal features encoding and fusion~\cite{liu2018attentive,liu2018cross,yuan2019find,chen2018temporally,jiang2019cross,chen2021drft,cao2021gtr},
or on 
generating high-quality candidate proposals~\cite{zhang2019man,zhang20202dtan,cao2020adversarial,xiao2021lpnet,zeng2021mmrg,liu2021cbln}. 
Although fully-supervised VMR methods 
have been shown effective on learning activity's boundary, 
they rely heavily on fine-grained manual temporal labels,
which are not only labour-intensive and time-consuming 
but also ambiguity-prone and sensitive to subjective bias~\cite{anne2017MCN}.

\noindent{\bf Weakly-supervised video moment retrieval}.
Different from the fully-supervised methods, 
there is no temporal boundary annotation being available 
for learning weakly-supervised approaches,
liberated from the demand for large-scale temporal labels.
Most existing weakly-supervised
methods~\cite{mithun2019tga,ma2020vlanet,tan2021logan,zhang2020ccl} 
were based on contrastive learning, 
which is minimising the distance in video level between the videos and their matching queries from the dataset 
and maximising that of videos and the queries of others. 
Besides, some approaches~\cite{duan2018wsdec,lin2020scn} tried to 
locate video moments by incorporating moment retrieval with sentence reconstruction.
Recently, CRM~\cite{huang2021CRM} proposed to mine the cross-sentence relationship by exploiting the video-level paragraph descriptions as a whole.
Regardless of their remarkable progress in the weakly-supervised setting, 
all of these approaches suffer from lacking fine-grained temporal annotations 
and overpass valuable temporal boundary labels available in existing datasets,
resulting in avoidable waste.

\noindent{\bf Domain adaptation}.
There are many works in domain adaptation, especially unsupervised domain adaptation, 
including variants of 
maximum mean discrepancies~\cite{gretton2012MMD,long2015learning,long2017deep}, 
adversarial learning~\cite{long2017conditional,liu2016coupled,bousmalis2017unsupervised} 
and transportable plan modelling~\cite{courty2016optimal,chen2018re,damodaran2018deepjdot} 
to measure and reduce the domain discrepancy.
However, currently, there are a few works focusing on deep domain adaptation 
in multi-modal applications.
Qi~\etal.~\cite{qi2018unified} proposed a unified framework for the general multi-modal task with an adaptive modal fusion and domain constraints.
Xu~\etal.~\cite{xu2019DAVQA} did domain adaptation on the video question answering task 
under a supervised domain adaptation setting. 
Chen~\etal.~\cite{chen2021mindgap} and Liu~\etal.~\cite{liu2021acp} proposed unsupervised domain adaptation methods for text-video retrieval from several trimmed video candidates. 
However, existing studies are mostly on knowledge adaptation across data domains while our objective is to adapt common video-text relationships across both domains and tasks with different vocabularies. Such a problem is both non-trivial and not straightforward to directly deploy existing methods.

\begin{figure}[t]
   \centering
   \includegraphics[width=\textwidth]{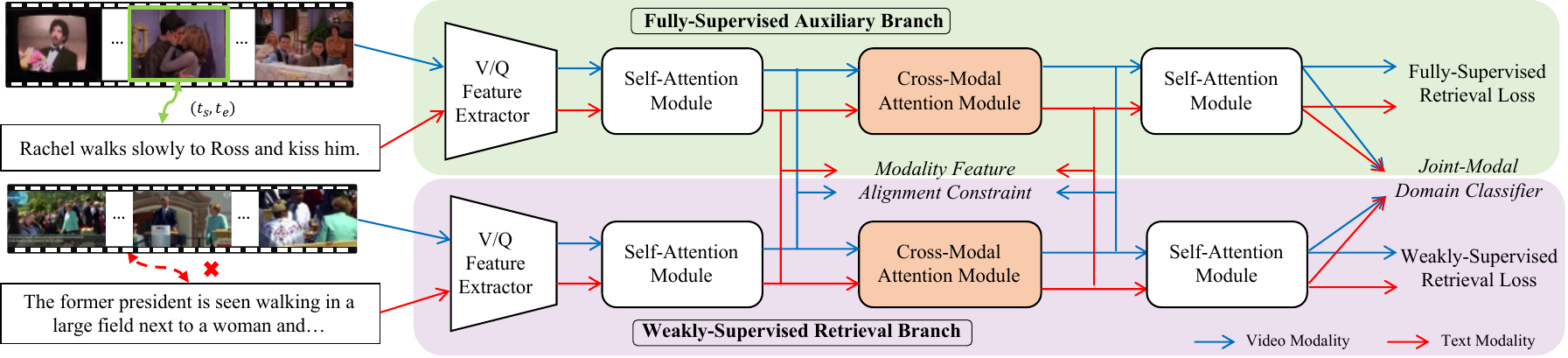}
   \vspace{-0.5cm}
   \caption{An overview of the proposed \textit{multiplE branch Video-text Alignment model} (\protect\method{abbr}) for hybrid-learning video moment retrieval.}
    \label{fig:framework}
\end{figure}

\section{Method}

We construct a
multiple branch network for model learning from a hybrid mixture
labelled training data in two different domains (Fig.~\ref{fig:framework}).
Suppose there 
is
an untrimmed video $V$ with $n_c$ consecutive
disjoint clips $V = \{c_i\} ^ {n_c} _ {i=1}$ in inflexible durations,
where $c_i$ is the $i$-th clip, and a corresponding query $Q$ with
$n_w$ words $Q= \{w_i\} ^ {n_w} _ {i=1}$, 
where $w_i$ is the
$i$-th word in the query sentence. 
In the
\textit{weakly-supervised retrieval branch}, 
videos $V^w$ and corresponding
queries $Q^w$ from the temporally-unlabelled (target) dataset
$D^w$ are given. 
In model training, the \textit{fully-supervised auxiliary branch} 
has the access to video-query pairs $(V^f, Q^f)$ 
with temporal labels $(t^{f}_s, t^{f}_e)$ 
from a fully-labelled auxiliary (source) dataset $D^f$. 
Our hybrid-learning VMR's goal
is to locate the most likely time boundaries $(\hat{t}^w_s,
\hat{t}^w_e)$ of the matching moment $\hat{S}^w$ to a given query $Q^w$ 
by learning to
optimise concurrently the pairing of both $(V^w, Q^w)$ and $(V^f, Q^f, t^{f}_s,
t^{f}_e)$. 
Temporal labels $(t^{f}_s, t^{f}_e)$ are 
mapped to their corresponding indices $(i^{f}_s, i^{f}_e)$ in video clips~$V^f$.
For clarity concern, we use 
$*^w$ and $*^f$ to distinguish symbols 
from the weakly-supervised retrieval branch (target domain) 
and fully-supervised auxiliary branch (source domain) 
when needed and omit them in formulations/modules applied in both domains.
For feature extraction,
we utilise a pre-trained 3D-CNN
model 
to acquire the clip-level video features with dimension $d_c$,
and GloVe word embeddings~\cite{pennington2014glove} 
as word-level
query features with dimension $d_w$. 
Both features are then projected into
$d$-dimensional spaces by two independent fully-connected layers.

\vspace{-0.2cm}
\subsection{Multi-Branch Hybrid-Learning}
\vspace{-0.1cm}
\noindent
The multi-branch hybrid-learning network
  is designed to augment
a \textit{weakly-supervised retrieval branch} 
on target domain 
unlabelled data
by an additional
\textit{fully-supervised auxiliary branch} given a
labelled auxiliary dataset.

\noindent\textbf{Weakly-supervised retrieval branch.} 
By exploring Transformer~\cite{vaswani2017transformer},
\method{abbr} employs several
attention modules to establish within- or cross-modality connections.
Given $Y\in \mathbb{R}^{l_y \times d}$ and $X\in
\mathbb{R}^{l_x \times d}$, the attention unit
$Att(Y,X)$ attends $Y$ using $X$ as follows: 
\vspace{-0.1cm}
\begin{equation}
\begin{gathered}
\mathcal{R}(Y,X) = \text{softmax}(Y{W^q}^\top W^k{X}^\top / \sqrt{d}), \quad
Att(Y,X) = \text{FC}(Y + \mathcal{R}(Y,X)X{W^v}^\top),
\end{gathered}
\label{eq:attention}
\vspace{-0.1cm}
\end{equation}
where the $\text{softmax}(\cdot)$ 
is the softmax normalisation 
by each row in the given matrix, 
and \{$W^q; W^k; W^v\} \in \mathbb{R}^{3 \times d \times d}$
are three trainable weight vectors.
A fully-connected layer $\text{FC}(\cdot)$ has the same dimension after projection.
Thus, the within-modal self-attention modules for video and text modalities are: 
\vspace{-0.1cm}
\begin{equation}
V \leftarrow Att^V(V, V), \quad
Q \leftarrow Att^Q(Q, Q),
\label{eq:self_attn}
\vspace{-0.1cm}
\end{equation}
which focus on 
exploring the within-modal dependencies
by learning the correlations between pairs of elements in 
a video or a text sentence.
We employ two sets of 
self-attention modules for video and text sequentially. 
Between them, 
the video is divided from clips 
into proposals $V\!=\!\{p_k\} ^ {n_p} _ {k=1}$ 
with a sliding window strategy~\cite{lin2020scn,ma2020vlanet}.
Then we fuse the query $Q
$ 
and proposal feature $p_k$
to construct a joint video-text representation
$j_{k}$~\cite{gao2017tall}
and 
acquire matching score $P_m(p_k \vert Q)$ 
by a fully connected layer 
$\text{FC}(\cdot)$ 
and a sigmoid function 
$ \sigma(\cdot)$:
\vspace{-0.1cm}
\begin{equation}
\begin{aligned}
P_m(p_k \vert Q)=\sigma(\text{FC}(j_{k})), 
\quad
j_{k} = (p_k + max(Q)) \Vert (p_k \otimes max(Q)) 
\Vert \text{FC}(p_k \Vert max(Q)),
\end{aligned}
\label{eq:score}
\vspace{-0.1cm}
\end{equation}
where $+$, $\otimes$, $\Vert$ are the element-wise matrix
addition, multiplication, and concatenation.  
Symbol $max(\cdot)$ denotes max-pooling to generate
sentence-level query features by aggregating all the words in a query.
The video-level matching score $P_m(V \vert Q)$ 
is obtained by the max-pooling of 
the proposal-level video-text matching score matrix 
$\{P_m(p_k \vert Q)\}^{n_p} _ {k=1}$. 
For each positive pair $(V,Q)$,
two negative counterparts $(V^-,Q)$ and $(V,Q^-)$ are sampled
by replacing either $V$ or $Q$ with $V^-$ or $Q^-$ from the mini-batch following~\cite{huang2021CRM}.
A binary cross-entropy (BCE) loss is then used for weakly-supervised
learning: 
\vspace{-0.1cm}
\begin{equation}
\begin{aligned}
\mathcal{L}_w 
= & 2 * (-\log P_m(V \vert Q))
-\log (1 - P_m(V \vert Q^-)) -\log (1 - P_m(V^- \vert Q)).
\end{aligned}
\label{eq:wloss}
\end{equation}

\noindent\textbf{Fully-supervised auxiliary branch.} 
In this branch, we want to proactively explore existing temporally labelled
video data from elsewhere to augment the weakly-supervised
learning in the target domain.
To project features to spaces with the same dimension as those in the
weakly-supervised branch, and to boost the intra-modal
contextual interaction in the source domain auxiliary dataset,
we use the same attention unit design
as in the weakly-supervised retrieval branch.
Per Eq.~(\ref{eq:attention}), 
a cross-modal attention module is deployed to 
learn the video-text interaction by 
focusing on the most matched parts between the two modality spaces: 
\vspace{-0.2cm}
\begin{equation}
V \leftarrow Att^{Q \rightarrow V}(V,Q), \quad
Q \leftarrow Att^{V \rightarrow Q}(Q,V),
\label{eq:cross_attn}
\vspace{-0cm}
\end{equation}
and shared parameters to weakly-supervised branch,
so as to promote more accurate
cross-modal interaction in weakly-supervised retrieval learning.
Then we compute video-query similarity score $S_{v-q} = V^fQ_m^f$, 
where $Q_m^f$ is the modularized query vector from $Q$~\cite{lei2020tvr,yu2018mattnet},
and use two 1D convolution filters to predict the start
and end boundary in the score curves, computed by 
$P_{s/e} = \sigma (\text{conv1D}_{s/e}(S_{v-q}))$, 
and $P_{s}$, $P_{e}$ denote the start and end probabilities. 
The fully-supervised retrieval loss is a weighted combination of a
retrieval loss $\mathcal{L}_{r}^f$ and the video-level BCE loss 
per Eq.~(\ref{eq:wloss}):
\vspace{-0.1cm}
\begin{equation}
\mathcal{L}_f = \lambda_{r}\mathcal{L}_{r}^f + \mathcal{L}_{bce}^f, \quad
\mathcal{L}_{r}^f
= H(P_{s}, I^{f}_s) + H(P_{e}, I^{f}_e),
\label{eq:floss}
\vspace{-0.1cm}
\end{equation}
where $H$ is the cross-entropy function,
$(I^{f}_s, I^{f}_e)$ are the one-hot ground-truth labels for start and end indices $(i^{f}_s, i^{f}_e)$,
and
$\lambda_{r}$ is a trade-off hyper-parameter.

\subsection{Multi-Modal Feature Alignment across Tasks and Domains}
As discussed earlier, 
there are apparent domain distribution shifts 
and data characteristic differences among heterogeneous datasets 
in both video and text modalities,
which bring challenges when utilising several datasets simultaneously.
Here our goal is to bridge the gaps across tasks and domains, 
and to transfer 
the video-text matching relationships 
learned from the fully-supervised source domain 
to the weakly-labelled target domain
effectively.

\subsubsection{Modality Feature Alignment Constraint}
Considering the differences in video and text modalities, diverse
datasets may represent the same semantic information in different
styles.
The apparent domain gaps in both natural language and videos among datasets
make any cross-modal interaction 
likely to be subject to
domain biases. 
This will introduce noise to the moment-level temporal labels given by the external fully-labelled dataset when shared with the weakly-supervised learning process in a different domain. 
For more accurate cross-modal matching 
and more effective precise label information sharing,
we propose to align the features
both {\em before} and {\em after} the cross-modal attention module in both modalities respectively. 
Specifically, 
by quantising and minimising the distance of the distributions between the source and target domains,
we constrain the learning of the source and target domain
to let the feature spaces for video and text modalities share
more common areas between the two domains. After aligning the
modality spaces on both input and output sides of the shared
cross-modal attention module,  
the video-text correlations learned on the fully-labelled auxiliary dataset
is more
generalisable and comprehensible to benefit the weakly-supervised
retrieval. 
Motivated by~\cite{xu2019DAVQA},
we use the maximum mean discrepancy (MMD)~\cite{gretton2012MMD} to measure the distribution's distance and constrain the intermediate features' distributions between source and target domain in both video and text modalities by minimising it.
Given source domain samples $D^s = \{ \bm{x}_i\}_{i=1}^{n_s} \in
\mathbb{R}^{n_s \times d} $ and target domain samples $D^t =\{
\bm{y}_j\}_{j=1}^{n_t} \in \mathbb{R}^{n_t \times d}$, the MMD is
calculated as  
: 
\vspace{-0.2cm}
\begin{equation}
\begin{aligned}
M(D^s, D^t)^2 = 
 \frac{1}{n_s^2}\sum_{i=1}^{n_s}\sum_{j=1}^{n_s}K(\bm{x}_i,\bm{x}_j) + 
\frac{1}{n_t^2}\sum_{i=1}^{n_t}\sum_{j=1}^{n_t}K(\bm{y}_i,\bm{y}_j)
 - \frac{2}{n_sn_t}\sum_{i=1}^{n_s}\sum_{j=1}^{n_t}K(\bm{x}_i,\bm{y}_j),
\end{aligned}
\vspace{-0.2cm}
\end{equation}
where $K$ is the radial basis function kernel~\cite{scholkopf2004RBFkernel}.
To align the modality features in both branches, we calculate
the MMD loss both before ($(V_b^f, V_b^w), (Q_b^f, Q_b^w)$) and after
($(V_a^f, V_a^w)$, $(Q_a^f, Q_a^w)$) 
the cross-modal attention module
between the fully-supervised auxiliary branch and the
weakly-supervised retrieval branch in visual and text modalities
respectively: 
\vspace{-0.1cm}
\begin{equation}
\begin{aligned}
\mathcal{L}_{align} = (\lambda_{vid}M(V_b^f, V_b^w)^2 + M(Q_b^f, Q_b^w)^2)
  + (\lambda_{vid}M(V_a^f, V_a^w)^2 + M(Q_a^f, Q_a^w)^2),
\end{aligned}
\label{eq:align}
\vspace{-0.1cm}
\end{equation}
where $\lambda_{vid}$ in Eq. (\ref{eq:align}) is a trade-off hyper-parameter between video and query modalities.

\subsubsection{Joint-Modal Domain Classifier}
Given that different modality features have their own
discriminative characteristics, we explore an adversarial strategy to
bridge the domain gaps whilst keeping per-task discriminativeness, with the
constraints of retrieval losses ($\mathcal{L}_w, \mathcal{L}_f$) in both branches. 
Here, we use a joint-modal domain classifier with the gradient reversal layer (GRL)~\cite{ganin2015DANN,qi2018unified} to tackle the problem.
The joint-modal domain classifier is designed to map the video feature $V$ and query feature $Q$ 
to a scalar domain label $label_d \in \{0,1\}$. The domain label is designed to distinguish whether the network input video and query are from the dataset with temporal boundaries (source domain) or the dataset without temporal boundaries (target domain).
The mapping can be expressed as $label_d = G_{d}(\cdot)$. 
Firstly, we concatenate the video feature $V$ and query feature $Q$ after max-pooling 
to generate a video-text joint feature $J$. 
And we apply another MMD constraint on the $V$ and $Q$.
Then $J$ is fed into 
the domain classifier $G_{d}(\cdot)$ 
which contains two fully-connected and a softmax layers:
\vspace{-0.1cm}
\begin{equation}
G_d(J) = \text{softmax}(\text{FC}_1(\text{FC}_2(J))).
\vspace{-0.1cm}
\end{equation}
For the purpose of optimising the model to get the features $V$ and $Q$ domain-invariant, we take the gradient reversal layer before the domain classifier $G_{d}(\cdot)$. 
The binary cross-entropy loss function is then adopted as 
the joint-modal domain classifier loss:
\vspace{-0.1cm}
\begin{equation}
\mathcal{L}_{domain} = -\text{log}(1-G_{d}(J^f))-\text{log}G_{d}(J^w),
\label{eq:domain}
\vspace{-0.1cm}
\end{equation}
where $J^f, J^w$ denote the joint features in the two branches respectively.

\subsection{Model Training and Testing}

In each training iteration, we randomly sample $n$ videos with a pair of queries from the target temporal-unlabelled dataset and the same amount samples with time annotations from the external temporal-labelled dataset, as a mini-batch. The overall loss is computed by:
\vspace{-0.2cm}
\begin{equation}
\mathcal{L} = \mathcal{L}_w +\lambda_{f}\mathcal{L}_f + \lambda_{align}\mathcal{L}_{align} - \lambda_{domain}\mathcal{L}_{domain},
\label{eq:loss}
\vspace{-0.2cm}
\end{equation}
where $\lambda_{f}, \lambda_{align}$ and $\lambda_{domain}$ are hyper-parameters for each loss.
In test, only the weakly-supervised retrieval branch is deployed.

\section{Experimental Results}

\vparagraph{Datasets.}

\begin{wraptable}{r}{0.56\linewidth}
\resizebox{\linewidth}{!}{%
\setlength{\tabcolsep}{0.04cm}
\begin{tabular}{l|c|c|c|c|c|c|c}
\hline
\multirow{2}{*}{Dataset} & \multirow{2}{*}{\#video} & \multicolumn{3}{c|}{\#moment} & \multicolumn{2}{c|}{avg. len. (sec)} & avg. len. (wrd) \\ \cline{3-8} 
 &  & train & val & test & video & moment & query \\ \hline \hline
Anet~\cite{krishna2017anetcaptions} 
& 19290 
& 37417 & 17505/17031 & - 
& 117.6 & 36.2 
& 14.8 \\ 
Charades~\cite{gao2017tall}
 & 6672 
 & 12408 & - & 3720 
 & 30.6 & 8.1 
 & 7.2 \\ 
TVR~\cite{lei2020tvr} 
& 21793 
& 87175 & 10895 & 5445 
& 76.2 & 9.1
& 13.4 \\ \hline
\end{tabular}
}
\vspace{-0.1cm}
\caption{Statistics of VMR datasets.}
\label{tab:dataset}
\end{wraptable}

In experiments, we employed three commonly used 
VMR datasets: ActivityNet-Captions~\cite{krishna2017anetcaptions,caba2015activitynet}, Charades-STA~\cite{gao2017tall} and newly released TVR~\cite{lei2020tvr}. 
The statistics of them are shown in Table~\ref{tab:dataset}.

For more extensive comparisons and to align with other weakly-supervised 
methods,
we use Charades and Anet
as the target datasets for weakly-supervised retrieval learning and
comparative evaluation. 
For hybrid learning, we use TVR for
the auxiliary training dataset with full temporal labelling, 
considering its large number of samples to cover greater linguistic diversity 
and precise video-text information, which is shown in Table~\ref{tab:dataset}.

\noindent\textbf{Evaluation protocol}.\quad 
Following prior works~\cite{duan2018wsdec,wu2020bar}, we use $IoU=m$, 
to calculate the percentage of the top predicted moment having 
Intersection over Union (IoU) larger than $m$.

\noindent\textbf{Implementation details}.\quad 
We used C3D features after PCA (500-D)
for per-frame representations in Anet,
I3D (1024-D) for Charades, 
and either C3D or I3D for TVR depending on the features used on the other domain. 
GloVe embeddings~\cite{pennington2014glove} 
were used as the word-level feature representations (300-D). 
The hidden features' dimension $d$ for 
both video clips and word representations were 256-D. 
For the weakly-supervised retrieval branch, 
the sliding windows stride was 8 and 
the window sizes were $\{8,16,32,64,128\}$ in Anet and
$\{8,12,20,32,64\}$ in Charades. 
The model was trained 50 epochs by Adam optimiser with a batch size of 64 
and learning rate of $1e-4$. 
The trade-off hyper-parameters were set as $\lambda_{r}^f = 0.1, \lambda_{f}=1, \lambda_{vid} = 0.8, \lambda_{domain} = 0.01, \lambda_{align}=1$.

\subsection{
Comparisons with the State-of-the-art
}

\noindent\textbf{Comparisons with fully-supervised methods.} 
The comparative evaluations on \method{abbr} in
Table~\ref{tab:cmp-f} are designed for a practical scenario 
which does not assume the target new data was drawn from the
identical distributions as the one used to train the model. 
Fully-supervised models rely heavily on manual temporal labels 
and lack the design to utilise or finetune on other datasets 
when temporal boundary labels are not available.
In this case, 
the fully-supervised methods were trained with the TVR dataset with
  full temporal annotations in model training, and evaluated on
Charades and Anet val\_2 respectively. 
The `Source' and `Target' columns indicate 
the access of temporal-labelled 
and unlabelled data in the training stage.
Table~\ref{tab:cmp-f} shows all fully-supervised methods 
trained on a specific dataset 
suffer a serious performance degradation
when deployed to a new domain in test, 
demonstrating their poor generalisation abilities. 
Our proposed 
\method{abbr} outperforms all these methods in all metrics.
Compared to \method{abbr}, all these fully-supervised methods are not designed for 
and cannot
simultaneously learn jointly from cross-domain hybrid labelling information
given by both fully labelled and weakly labelled data 
in different domains.
Our new multi-branch hybrid learning model
demonstrates compellingly its ability to utilise and exploit effectively
cross-domain different training labels.

\begin{table}[t]
\centering
\resizebox{0.7\columnwidth}{!}{%
\begin{tabular}{l|c|c|c|c|c|c|c|c}
\hline
\multirow{2}{*}{Method} & 
\multirow{2}{*}{Source} & 
\multirow{2}{*}{Target} & 
\multicolumn{3}{c|}{Charades} & 
\multicolumn{3}{c}{Anet} \\ \cline{4-9} 
 &  &  
 & IoU=0.3 & IoU=0.5 & IoU=0.7 
 & IoU=0.1 & IoU=0.3 & IoU=0.5 \\ \hline \hline
2D-TAN~\cite{zhang20202dtan} 
& \cmark & \xmark 
& 14.65 & 4.30 & 1.26 
& 40.16 & 28.71 & 17.29 \\
XML~\cite{lei2020tvr} 
& \cmark & \xmark 
& 32.49 & 18.27 & 8.87 
& 31.78 & 17.18 & 9.27 \\
MMN~\cite{wang2021mmn} 
& \cmark & \xmark  
& 11.45 & 3.06 & 0.86 
& 42.19 & 24.57 & 13.09 \\ \hline
\method{abbr} 
& \cmark & \cmark
& \textbf{62.01} & \textbf{40.21} & \textbf{18.22} 
& \textbf{74.09} & \textbf{49.89} & \textbf{29.43} \\ \hline
\end{tabular}%
}
\vspace{0.3cm}
\caption{Performance comparisons 
between \protect\method{abbr} hybrid-learning and 
state-of-the-art fully-supervised VMR methods tested on Charades
and Anet. 
}
\label{tab:cmp-f}
\end{table}

\begin{table}[t]
\centering
\begin{subtable}[h]{0.38\textwidth}
\centering
\setlength{\tabcolsep}{0.05cm}
\resizebox{\linewidth}{!}{%
\begin{tabular}{l|c|c|c|c|c}
\hline
\multirow{1}{*}{Method} & 
\multirow{1}{*}{Source} & 
\multirow{1}{*}{Target} & 
\multicolumn{1}{c|}{IoU=0.3} & 
\multicolumn{1}{c|}{IoU=0.5} & IoU=0.7 \\ \hline \hline
TGA~\cite{mithun2019tga}
& \xmark & \cmark 
& 29.68 & 17.04 & 6.93 \\
SCN~\cite{lin2020scn}
& \xmark & \cmark 
& 42.96 & 23.58 & 9.97 \\
LoGAN~\cite{tan2021logan}
& \xmark & \cmark  
& 51.67 & 34.68 & 14.54 \\
BAR~\cite{wu2020bar}
& \xmark & \cmark 
& 44.97 & 27.04 & 12.23 \\
RTBPN~\cite{zhang2020rtbpn}
& \xmark & \cmark 
& 60.04 & 32.36 & 13.24 \\
VLANet~\cite{ma2020vlanet}
& \xmark & \cmark  
& 45.24 & 31.83 & 14.17 \\
CCL~\cite{zhang2020ccl}
& \xmark & \cmark 
& - & 33.21 & 15.68 \\
CRM~\cite{huang2021CRM}
& \xmark & \cmark 
& 53.66 & 34.76 & 16.37 \\ \hline
\method{abbr} 
& \cmark & \cmark 
& \textbf{62.01} & \textbf{40.21} & \textbf{18.22} \\ \hline
\end{tabular}
}
\vspace{-0.15cm}
\caption{Evaluated on Charades}
\end{subtable}
\hfil
\begin{subtable}[h]{0.38\textwidth}
\centering
\setlength{\tabcolsep}{0.05cm}
\resizebox{\linewidth}{!}{%
\begin{tabular}{l|c|c|c|c|c|c}
\hline	
\multirow{1}{*}{Method} & 
\multirow{1}{*}{Source} & 
\multirow{1}{*}{Target} & 
\multirow{1}{*}{Split} & 
\multicolumn{1}{c|}{IoU=0.1} & 
\multicolumn{1}{c|}{IoU=0.3} & IoU=0.5 \\ \hline \hline
WS-DEC~\cite{duan2018wsdec}
& \xmark & \cmark 
& val\_1
& 62.71 & 41.98 & 23.34 \\
WSLLN~\cite{gao2019wslln}
& \xmark & \cmark 
& val\_1
& 75.4 & 42.8 & 22.7 \\
BAR~\cite{wu2020bar}
& \xmark & \cmark 
& val\_1
& - & 49.03 & 30.73 \\
CRM~\cite{huang2021CRM}
& \xmark & \cmark 
& val\_1
& \textbf{76.66} & \textbf{51.17} & \textbf{31.67} \\ \hline
\method{abbr} %
& \cmark & \cmark 
& val\_1
& 70.79 & 46.23 & 28.00 \\ \hline \hline %
SCN~\cite{lin2020scn}
& \xmark & \cmark 
& val\_2
& 71.48 & 47.23 & 29.22 \\
RTBPN~\cite{zhang2020rtbpn}
& \xmark & \cmark 
& val\_2
& 73.73 & 49.77 & 29.63 \\
CCL~\cite{zhang2020ccl}
& \xmark & \cmark 
& val\_2
& - & 50.12 & 31.07 \\
CRM~\cite{huang2021CRM}
& \xmark & \cmark 
& val\_2
& \textbf{81.61} & \textbf{55.26} & \textbf{32.19} \\ \hline
\method{abbr} 
& \cmark & \cmark 
& val\_2
& 74.09 & 49.89 & 29.43 \\ \hline
\end{tabular}%
}
\vspace{-0.15cm}
\caption{Evaluated on Anet}
\end{subtable}
\vspace{0.2cm}
\caption{Comparisons with state-of-the-art weakly-supervised VMR methods.}
\label{tab:cmp-w}
\end{table}

\noindent\textbf{Comparisons with weakly-supervised approaches.} 
Table \ref{tab:cmp-w} compares EVA with the state-of-the-art
weakly-supervised models. All these methods 
have no access to temporal labels in training.  
It is evident that \method{abbr} performs well on
Charades which shows the ability to introduce precise video-text
interaction information from an external heterogeneous
temporal-labelled dataset.  
In Anet, some
powerful weakly-supervised methods will maintain better performance.
Our analysis is that some of these methods
are likely to have overfitted a training dataset.
The video-query pairs from prevailing
datasets suffer annotation biases~\cite{yuan2021ood}, where for
instance the moments will fall into several specific time locations in
both train and test splits. 
Due to this, a model trained with such a dataset 
may only make retrieval by selecting a moment from the
frequency statistics of the bias-based training split and still have
satisfactory performance on the test set which shares similarly biased
distribution.

\begin{wraptable}{r}{0.55\linewidth}
\centering
\resizebox{\linewidth}{!}{%
\begin{tabular}{l|l|c|c|c|c|c}
\hline
Dataset & Method & Source & Target 
& IoU=0.3 & IoU=0.5 & IoU=0.7 \\ \hline \hline
\multirow{3}{*}{Anet} 
& WS-DEC~\cite{duan2018wsdec}
& \xmark & \cmark
& 17.00 & 7.17 & 1.82 \\ 
& CRM~\cite{huang2021CRM}
& \xmark & \cmark
& 22.77 & 10.31 & - \\ 
\cline{2-7} 
& \method{abbr} 
& \cmark & \cmark 
& \textbf{23.11} & \textbf{11.29} & \textbf{4.32} \\ 
\hline \hline 
\multirow{2}{*}{Charades} 
& WS-DEC~\cite{duan2018wsdec}
& \xmark & \cmark
& 35.86 & 23.67 & 8.27 \\ \cline{2-7} 
& \method{abbr} 
& \cmark & \cmark
& \textbf{47.83} & \textbf{31.71} & \textbf{12.76} \\ \hline
\end{tabular}
}
\vspace{0cm}
\caption{\protect\method{abbr} hybrid-learning VMR results on
    ActivityNet-CD and Charades-CD OOD splits. 
    } 
\label{tab:cmp-ood}
\end{wraptable}

To test our assumption on weakly-supervised model overfitting, 
we further carried out experiments 
on distribution changed splits for
ActivityNet-CD and 
Charades-CD~\cite{yuan2021ood} and evaluated on the
out-of-distribution (OOD) test splits. Table \ref{tab:cmp-ood}
shows the results under the discounted recall
metric~\cite{yuan2021ood},
comparing \method{abbr} 
with the best performing state-of-the-art weakly-supervised models
WS-DEC and CRM. 
It is evident that EVA outperforms both methods, 
showing that \method{abbr} has 
great multi-modal understanding by exploiting and sharing the precise
video-text interaction information from an external dataset of
different labels in a different domain context.

\subsection{
Discussion and Analysis
}

\noindent\textbf{Ablation study.}
We examined the effectiveness of each proposed component in
Table \ref{tab:abla-comp}. 
`WR' and `FA' are abbreviations of the weakly-supervised retrieval 
and fully-supervised auxiliary branches, 
whilst the `Align' and `Domain' mean the modality feature alignment constraint and joint-modal domain classifier.
In the `WR + fine-tune' row, we pre-trained the WR branch on the source dataset with temporal labels and fine-tuned it on the target dataset without labels,
and in `WR + unlabelled source', the WR branch was trained jointly without any temporal labels in both the source and target domain training data.
Our WR branch was
trained with a temporal-unlabelled target dataset only as of the baseline, 
and all other mentioned methods have the access to both
target and source datasets. 
The results show that simply having an external temporal-labelled dataset 
gains limited improvement in some
metrics but not all. Critically, the performance is limited
due to the multi-modal domain gaps.  
The modality feature alignment constraint we introduced in each
modality and the joint-modal domain classifier proposed in \method{abbr} to align the
single-modal and joint-modal features respectively not only have
their own benefits, but also when they are both adopted
together by aligning the features both in each modality and
  cross-modality, the model performance benefited more.

\noindent\textbf{Effect of module sharing.}
To promote precise cross-modal matching information interaction in two branches, 
we shared the parameters of the cross-modal attention modules
and maintain the independence of self-attention modules to focusing
on each domain intra-modal interaction. 
We investigated its effect by
comparing the prediction recall of~\method{abbr} constructed with
different sharing parts on Anet val\_2 split 
in Table~\ref{tab:abla-share}, displaying its advantages in video-text
knowledge sharing.  
`Self1', `Self2', and `Cross' refer to the first, second self-attention module, and cross-modal attention module respectively.

\begin{table}[t]
\centering
\setlength{\tabcolsep}{0.1cm}
\begin{minipage}{0.46\linewidth}\centering
\resizebox{\linewidth}{!}{%
\begin{tabular}{l|c|c|c|c}
\hline
\multirow{1}{*}{Method} 
 & mIoU & IoU=0.1 & IoU=0.3 & IoU=0.5 \\ \hline \hline
WR 
& 32.96	& 71.48	& 48.06	& 28.74 \\
WR + fine-tune
& 33.14	& 72.59	& 48.86	& 28.21 \\
WR + unlabelled source
& 33.01	& 72.18	& 48.18	& 28.01 \\
WR + FA
& 33.11	& 71.62	& 48.46	& 28.56 \\ 
WR + FA + Align 
& 33.83	& 73.35	& 49.60	& 28.80 \\ 
WR + FA + Domain
& 33.66	& 73.14	& 49.32	& 28.92 \\ \hline
WR + FA + Align + Domain
& \textbf{34.27} & \textbf{74.09} & \textbf{49.89} & \textbf{29.43} \\ \hline
\end{tabular}
}
\vspace{0cm}
\caption{Component ablation study of \protect\method{abbr} on TVR (source) and Anet (target). 
}
\label{tab:abla-comp}
\end{minipage}
\hfil
\begin{minipage}{0.43\linewidth}\centering
\resizebox{\linewidth}{!}{%
\begin{tabular}{l|c|c|c|c}
\hline
Shared Module(s) & mIoU & IoU=0.1 & IoU=0.3 & IoU=0.5 \\ \hline \hline 
No Sharing 
& 33.85	& 73.40	& 49.73	& 29.21  \\ 
Self1 + Self2 
& 34.08	& 73.63	& 49.84	& 29.33\\ 
Self1 + Cross 
& 34.00 & 73.92	& 49.21	& 28.91 \\ 
Self2 + Cross 
& 33.88	& 73.81	& 49.42	& 28.80  \\ 
Self1 + Self2 + Cross 
& 33.70	& 73.92	& 49.06	& 28.30  \\ \hline
Cross & \textbf{34.27} & \textbf{74.09} & \textbf{49.89} & \textbf{29.43} \\ \hline
\end{tabular}
}
\vspace{0cm}
\caption{Effects of module sharing. 
}
\label{tab:abla-share}
\end{minipage}
\end{table}

\begin{table}[t]
\centering
\resizebox{0.8\columnwidth}{!}{%
\begin{tabular}{lcc|cccc|cccc}
\hline
\multicolumn{1}{l|}{\multirow{2}{*}{Model}} & \multicolumn{2}{c|}{Train Dataset} & \multicolumn{4}{c|}{(t,a)$\rightarrow$c} & \multicolumn{4}{c}{(t,a)$\rightarrow$a} \\ \cline{2-11} 
\multicolumn{1}{l|}{} & \multicolumn{1}{c|}{Source} & Target & \multicolumn{1}{c|}{mIoU} & \multicolumn{1}{c|}{IoU=0.3} & \multicolumn{1}{c|}{IoU=0.5} & IoU=0.7 & \multicolumn{1}{c|}{mIoU} & \multicolumn{1}{c|}{IoU=0.1} & \multicolumn{1}{c|}{IoU=0.3} & IoU=0.5 \\ \hline
\multicolumn{1}{l|}{WR} 
& \multicolumn{1}{c|}{\xmark} & \cmark
& \multicolumn{1}{c|}{22.68} & \multicolumn{1}{c|}{34.48} & \multicolumn{1}{c|}{17.84} & 6.23 
& \multicolumn{1}{c|}{31.50} & \multicolumn{1}{c|}{69.62} & \multicolumn{1}{c|}{46.05} & 26.23 \\
\multicolumn{1}{l|}{\method{abbr}} 
& \multicolumn{1}{c|}{\cmark} & \cmark
& \multicolumn{1}{c|}{\textbf{24.18}} & \multicolumn{1}{c|}{\textbf{36.91}} 
& \multicolumn{1}{c|}{\textbf{22.53}} & \textbf{8.79} 
& \multicolumn{1}{c|}{\textbf{34.04}} & \multicolumn{1}{c|}{\textbf{73.76}} 
& \multicolumn{1}{c|}{\textbf{49.80}} & \textbf{28.75} 
\\ \hline
 &  &  & \multicolumn{4}{c|}{(t,c)$\rightarrow$c} & \multicolumn{4}{c}{(t,c)$\rightarrow$a} \\ \cline{4-11} 
 &  &  & \multicolumn{1}{c|}{mIoU} & \multicolumn{1}{c|}{IoU=0.3} & \multicolumn{1}{c|}{IoU=0.5} & IoU=0.7 & \multicolumn{1}{c|}{mIoU} & \multicolumn{1}{c|}{IoU=0.1} & \multicolumn{1}{c|}{IoU=0.3} & IoU=0.5 \\ \hline
\multicolumn{1}{l|}{WR} 
& \multicolumn{1}{c|}{\xmark} & \cmark 
& \multicolumn{1}{c|}{39.74} & \multicolumn{1}{c|}{61.72} & \multicolumn{1}{c|}{38.60} & 16.63 
& \multicolumn{1}{c|}{17.94} & \multicolumn{1}{c|}{45.04} & \multicolumn{1}{c|}{24.47} & 12.35 \\
\multicolumn{1}{l|}{\method{abbr}} 
& \multicolumn{1}{c|}{\cmark} & \cmark
& \multicolumn{1}{c|}{\textbf{40.08}} & \multicolumn{1}{c|}{\textbf{62.01}} 
& \multicolumn{1}{c|}{\textbf{40.21}} & \textbf{18.22} 
& \multicolumn{1}{c|}{\textbf{22.04}} & \multicolumn{1}{c|}{\textbf{52.05}} 
& \multicolumn{1}{c|}{\textbf{31.81}} & \textbf{16.57} \\ \hline
\end{tabular}%
}
\vspace{0.4cm}
\caption{Model generalisation evaluation on \protect\method{abbr} hybrid-learning
that utilises different datasets in training and test.} 
\label{tab:abla-gen}
\end{table}

\noindent\textbf{Generalisation to unseen data.}
To explore the generalisation ability of \method{abbr}, we try
to use different datasets in training and evaluation stages. 
Specifically, 
we trained \method{abbr} on TVR,
Anet and tested on Charades ((t,a)$\rightarrow$c),
and then trained on TVR, Charades and tested on
Anet val\_2 ((t,c)$\rightarrow$a). 
All the videos are processed by I3D to extract the features. The results in
Table \ref{tab:abla-gen} show that \method{abbr} not only has
improvements on the target dataset, but also improves performance in all metrics
over the baseline (WR) on a heterogeneous dataset of a
different domain, which reveals that bringing 
an extra temporal-labelled dataset in our method can largely avoid the model from converging to
some bias-based distributions and carry more useful cross-modal
information for more precise moment localisation.

\section{
Conclusion}
In this work, we introduced a new hybrid-learning approach
 to VMR by formulating a novel multiple branch
video-text alignment framework. 
\method{abbr} explores fine-grained
cross-modal matching interaction information with a shared cross-modal
attention module between two branches. \method{abbr} also employs a
modality feature alignment constraint and joint-modal domain
classifier to align the features in individual and multiple modalities
so to preserve their per-task
discriminativeness. 
Experiments 
show the advantages of 
\method{abbr} over existing methods and demonstrate the effectiveness
of our hybrid learning model in improving
cross-domain weakly-supervised learning.

\section*{Acknowledgements}
This work was supported by the China Scholarship Council, Vision Semantics Limited, and the Alan Turing Institute Turing Fellowship.

\bibliography{vmr}

\begin{thebibliography}{50}
\providecommand{\natexlab}[1]{#1}
\providecommand{\url}[1]{\texttt{#1}}
\expandafter\ifx\csname urlstyle\endcsname\relax
  \providecommand{\doi}[1]{doi: #1}\else
  \providecommand{\doi}{doi: \begingroup \urlstyle{rm}\Url}\fi

\bibitem[Anne~Hendricks et~al.(2017)Anne~Hendricks, Wang, Shechtman, Sivic, Darrell, and Russell]{anne2017MCN}
Lisa Anne~Hendricks, Oliver Wang, Eli Shechtman, Josef Sivic, Trevor Darrell, and Bryan Russell.
\newblock Localizing moments in video with natural language.
\newblock In \emph{Proceedings of the IEEE International Conference on Computer Vision}, pages 5803--5812, 2017.

\bibitem[Bao and Mu(2022)]{bao2022crossscene}
Peijun Bao and Yadong Mu.
\newblock Learning sample importance for cross-scenario video temporal grounding.
\newblock \emph{arXiv preprint arXiv:2201.02848}, 2022.

\bibitem[Bousmalis et~al.(2017)Bousmalis, Silberman, Dohan, Erhan, and Krishnan]{bousmalis2017unsupervised}
Konstantinos Bousmalis, Nathan Silberman, David Dohan, Dumitru Erhan, and Dilip Krishnan.
\newblock Unsupervised pixel-level domain adaptation with generative adversarial networks.
\newblock In \emph{Proceedings of the IEEE Conference on Computer Vision and Pattern Recognition}, pages 3722--3731, 2017.

\bibitem[Caba~Heilbron et~al.(2015)Caba~Heilbron, Escorcia, Ghanem, and Carlos~Niebles]{caba2015activitynet}
Fabian Caba~Heilbron, Victor Escorcia, Bernard Ghanem, and Juan Carlos~Niebles.
\newblock Activitynet: A large-scale video benchmark for human activity understanding.
\newblock In \emph{Proceedings of the IEEE Conference on Computer Vision and Pattern Recognition}, pages 961--970, 2015.

\bibitem[Cao et~al.(2020)Cao, Zeng, Wei, Nie, Hong, and Qin]{cao2020adversarial}
Da~Cao, Yawen Zeng, Xiaochi Wei, Liqiang Nie, Richang Hong, and Zheng Qin.
\newblock Adversarial video moment retrieval by jointly modeling ranking and localization.
\newblock In \emph{Proceedings of the 28th ACM International Conference on Multimedia}, pages 898--906, 2020.

\bibitem[Cao et~al.(2021)Cao, Chen, Shou, Zhang, and Zou]{cao2021gtr}
Meng Cao, Long Chen, Mike~Zheng Shou, Can Zhang, and Yuexian Zou.
\newblock On pursuit of designing multi-modal transformer for video grounding.
\newblock \emph{arXiv preprint arXiv:2109.06085}, 2021.

\bibitem[Chen et~al.(2018{\natexlab{a}})Chen, Chen, Ma, Jie, and Chua]{chen2018temporally}
Jingyuan Chen, Xinpeng Chen, Lin Ma, Zequn Jie, and Tat-Seng Chua.
\newblock Temporally grounding natural sentence in video.
\newblock In \emph{Proceedings of the 2018 Conference on Empirical Methods in Natural Language Processing}, pages 162--171, 2018{\natexlab{a}}.

\bibitem[Chen et~al.(2018{\natexlab{b}})Chen, Liu, Wang, Wassell, and Chetty]{chen2018re}
Qingchao Chen, Yang Liu, Zhaowen Wang, Ian Wassell, and Kevin Chetty.
\newblock Re-weighted adversarial adaptation network for unsupervised domain adaptation.
\newblock In \emph{Proceedings of the IEEE Conference on Computer Vision and Pattern Recognition}, pages 7976--7985, 2018{\natexlab{b}}.

\bibitem[Chen et~al.(2021{\natexlab{a}})Chen, Liu, and Albanie]{chen2021mindgap}
Qingchao Chen, Yang Liu, and Samuel Albanie.
\newblock Mind-the-gap! unsupervised domain adaptation for text-video retrieval.
\newblock In \emph{Proceedings of the AAAI Conference on Artificial Intelligence}, volume~35, pages 1072--1080, 2021{\natexlab{a}}.

\bibitem[Chen et~al.(2021{\natexlab{b}})Chen, Tsai, and Yang]{chen2021drft}
Yi-Wen Chen, Yi-Hsuan Tsai, and Ming-Hsuan Yang.
\newblock End-to-end multi-modal video temporal grounding.
\newblock \emph{Advances in Neural Information Processing Systems}, 34:\penalty0 28442--28453, 2021{\natexlab{b}}.

\bibitem[Courty et~al.(2016)Courty, Flamary, Tuia, and Rakotomamonjy]{courty2016optimal}
Nicolas Courty, R{\'e}mi Flamary, Devis Tuia, and Alain Rakotomamonjy.
\newblock Optimal transport for domain adaptation.
\newblock \emph{IEEE transactions on Pattern Analysis and Machine Intelligence}, 39\penalty0 (9):\penalty0 1853--1865, 2016.

\bibitem[Damodaran et~al.(2018)Damodaran, Kellenberger, Flamary, Tuia, and Courty]{damodaran2018deepjdot}
Bharath~Bhushan Damodaran, Benjamin Kellenberger, R{\'e}mi Flamary, Devis Tuia, and Nicolas Courty.
\newblock Deepjdot: Deep joint distribution optimal transport for unsupervised domain adaptation.
\newblock In \emph{Proceedings of the European Conference on Computer Vision}, pages 447--463, 2018.

\bibitem[Duan et~al.(2018)Duan, Huang, Gan, Wang, Zhu, and Huang]{duan2018wsdec}
Xuguang Duan, Wenbing Huang, Chuang Gan, Jingdong Wang, Wenwu Zhu, and Junzhou Huang.
\newblock Weakly supervised dense event captioning in videos.
\newblock In S.~Bengio, H.~Wallach, H.~Larochelle, K.~Grauman, N.~Cesa-Bianchi, and R.~Garnett, editors, \emph{Advances in Neural Information Processing Systems}, volume~31, pages 3062--3072. Curran Associates, Inc., 2018.

\bibitem[Ganin and Lempitsky(2015)]{ganin2015DANN}
Yaroslav Ganin and Victor Lempitsky.
\newblock Unsupervised domain adaptation by backpropagation.
\newblock In \emph{Proceedings of the International Conference on Machine Learning}, pages 1180--1189. PMLR, 2015.

\bibitem[Gao et~al.(2017)Gao, Sun, Yang, and Nevatia]{gao2017tall}
Jiyang Gao, Chen Sun, Zhenheng Yang, and Ram Nevatia.
\newblock Tall: Temporal activity localization via language query.
\newblock In \emph{Proceedings of the IEEE International Conference on Computer Vision}, pages 5267--5275, 2017.

\bibitem[Gao et~al.(2019)Gao, Davis, Socher, and Xiong]{gao2019wslln}
Mingfei Gao, Larry~S Davis, Richard Socher, and Caiming Xiong.
\newblock Wslln: Weakly supervised natural language localization networks.
\newblock \emph{arXiv preprint arXiv:1909.00239}, 2019.

\bibitem[Gretton et~al.(2012)Gretton, Borgwardt, Rasch, Sch{\"o}lkopf, and Smola]{gretton2012MMD}
Arthur Gretton, Karsten~M Borgwardt, Malte~J Rasch, Bernhard Sch{\"o}lkopf, and Alexander Smola.
\newblock A kernel two-sample test.
\newblock \emph{The Journal of Machine Learning Research}, 13\penalty0 (1):\penalty0 723--773, 2012.

\bibitem[Huang et~al.(2021)Huang, Liu, Gong, and Jin]{huang2021CRM}
Jiabo Huang, Yang Liu, Shaogang Gong, and Hailin Jin.
\newblock Cross-sentence temporal and semantic relations in video activity localisation.
\newblock In \emph{Proceedings of the IEEE/CVF International Conference on Computer Vision}, pages 7199--7208, October 2021.

\bibitem[Jiang et~al.(2019)Jiang, Huang, Yang, and Yuan]{jiang2019cross}
Bin Jiang, Xin Huang, Chao Yang, and Junsong Yuan.
\newblock Cross-modal video moment retrieval with spatial and language-temporal attention.
\newblock In \emph{Proceedings of the 2019 on International Conference on Multimedia Retrieval}, pages 217--225, 2019.

\bibitem[Krishna et~al.(2017)Krishna, Hata, Ren, Fei-Fei, and Carlos~Niebles]{krishna2017anetcaptions}
Ranjay Krishna, Kenji Hata, Frederic Ren, Li~Fei-Fei, and Juan Carlos~Niebles.
\newblock Dense-captioning events in videos.
\newblock In \emph{Proceedings of the IEEE International Conference on Computer Vision}, pages 706--715, 2017.

\bibitem[Lei et~al.(2020)Lei, Yu, Berg, and Bansal]{lei2020tvr}
Jie Lei, Licheng Yu, Tamara~L Berg, and Mohit Bansal.
\newblock Tvr: A large-scale dataset for video-subtitle moment retrieval.
\newblock In \emph{Proceedings of the European Conference on Computer Vision}, pages 447--463. Springer, 2020.

\bibitem[Lin et~al.(2020)Lin, Zhao, Zhang, Wang, and Liu]{lin2020scn}
Zhijie Lin, Zhou Zhao, Zhu Zhang, Qi~Wang, and Huasheng Liu.
\newblock Weakly-supervised video moment retrieval via semantic completion network.
\newblock In \emph{Proceedings of the AAAI Conference on Artificial Intelligence}, volume~34, pages 11539--11546, 2020.

\bibitem[Liu et~al.(2021{\natexlab{a}})Liu, Qu, Dong, Zhou, Cheng, Wei, Xu, and Xie]{liu2021cbln}
Daizong Liu, Xiaoye Qu, Jianfeng Dong, Pan Zhou, Yu~Cheng, Wei Wei, Zichuan Xu, and Yulai Xie.
\newblock Context-aware biaffine localizing network for temporal sentence grounding.
\newblock In \emph{Proceedings of the IEEE/CVF Conference on Computer Vision and Pattern Recognition}, pages 11235--11244, 2021{\natexlab{a}}.

\bibitem[Liu et~al.(2018{\natexlab{a}})Liu, Wang, Nie, He, Chen, and Chua]{liu2018attentive}
Meng Liu, Xiang Wang, Liqiang Nie, Xiangnan He, Baoquan Chen, and Tat-Seng Chua.
\newblock Attentive moment retrieval in videos.
\newblock In \emph{The 41st International ACM SIGIR Conference on Research \& Fevelopment in Information Retrieval}, pages 15--24, 2018{\natexlab{a}}.

\bibitem[Liu et~al.(2018{\natexlab{b}})Liu, Wang, Nie, Tian, Chen, and Chua]{liu2018cross}
Meng Liu, Xiang Wang, Liqiang Nie, Qi~Tian, Baoquan Chen, and Tat-Seng Chua.
\newblock Cross-modal moment localization in videos.
\newblock In \emph{Proceedings of the 26th ACM International Conference on Multimedia}, pages 843--851, 2018{\natexlab{b}}.

\bibitem[Liu and Tuzel(2016)]{liu2016coupled}
Ming-Yu Liu and Oncel Tuzel.
\newblock Coupled generative adversarial networks.
\newblock \emph{Advances in Neural Information Processing Systems}, 29:\penalty0 469--477, 2016.

\bibitem[Liu et~al.(2021{\natexlab{b}})Liu, Chen, and Albanie]{liu2021acp}
Yang Liu, Qingchao Chen, and Samuel Albanie.
\newblock Adaptive cross-modal prototypes for cross-domain visual-language retrieval.
\newblock In \emph{Proceedings of the IEEE/CVF Conference on Computer Vision and Pattern Recognition}, pages 14954--14964, June 2021{\natexlab{b}}.

\bibitem[Long et~al.(2015)Long, Cao, Wang, and Jordan]{long2015learning}
Mingsheng Long, Yue Cao, Jianmin Wang, and Michael Jordan.
\newblock Learning transferable features with deep adaptation networks.
\newblock In \emph{Proceedings of the International Conference on Machine Learning}, pages 97--105. PMLR, 2015.

\bibitem[Long et~al.(2017{\natexlab{a}})Long, Cao, Wang, and Jordan]{long2017conditional}
Mingsheng Long, Zhangjie Cao, Jianmin Wang, and Michael~I Jordan.
\newblock Conditional adversarial domain adaptation.
\newblock \emph{arXiv preprint arXiv:1705.10667}, 2017{\natexlab{a}}.

\bibitem[Long et~al.(2017{\natexlab{b}})Long, Zhu, Wang, and Jordan]{long2017deep}
Mingsheng Long, Han Zhu, Jianmin Wang, and Michael~I Jordan.
\newblock Deep transfer learning with joint adaptation networks.
\newblock In \emph{Proceedings of the International Conference on Machine Learning}, pages 2208--2217. PMLR, 2017{\natexlab{b}}.

\bibitem[Ma et~al.(2020)Ma, Yoon, Kim, Lee, Kang, and Yoo]{ma2020vlanet}
Minuk Ma, Sunjae Yoon, Junyeong Kim, Youngjoon Lee, Sunghun Kang, and Chang~D Yoo.
\newblock Vlanet: Video-language alignment network for weakly-supervised video moment retrieval.
\newblock In \emph{Proceedings of the European Conference on Computer Vision}, pages 156--171. Springer, 2020.

\bibitem[Mithun et~al.(2019)Mithun, Paul, and Roy-Chowdhury]{mithun2019tga}
Niluthpol~Chowdhury Mithun, Sujoy Paul, and Amit~K Roy-Chowdhury.
\newblock Weakly supervised video moment retrieval from text queries.
\newblock In \emph{Proceedings of the IEEE/CVF Conference on Computer Vision and Pattern Recognition}, pages 11592--11601, 2019.

\bibitem[Pennington et~al.(2014)Pennington, Socher, and Manning]{pennington2014glove}
Jeffrey Pennington, Richard Socher, and Christopher~D Manning.
\newblock Glove: Global vectors for word representation.
\newblock In \emph{Proceedings of the 2014 Conference on Empirical Methods in Natural Language Processing}, pages 1532--1543, 2014.

\bibitem[Qi et~al.(2018)Qi, Yang, and Xu]{qi2018unified}
Fan Qi, Xiaoshan Yang, and Changsheng Xu.
\newblock A unified framework for multimodal domain adaptation.
\newblock In \emph{Proceedings of the 26th ACM International Conference on Multimedia}, pages 429--437, 2018.

\bibitem[Sch{\"o}lkopf et~al.(2004)Sch{\"o}lkopf, Tsuda, and Vert]{scholkopf2004RBFkernel}
Bernhard Sch{\"o}lkopf, Koji Tsuda, and Jean-Philippe Vert.
\newblock \emph{Kernel methods in computational biology}.
\newblock MIT press, 2004.

\bibitem[Song et~al.(2020)Song, Wang, Ma, Yu, and Yu]{song2020MARN}
Yijun Song, Jingwen Wang, Lin Ma, Zhou Yu, and Jun Yu.
\newblock Weakly-supervised multi-level attentional reconstruction network for grounding textual queries in videos.
\newblock \emph{arXiv preprint arXiv:2003.07048}, 2020.

\bibitem[Tan et~al.(2021)Tan, Xu, Saenko, and Plummer]{tan2021logan}
Reuben Tan, Huijuan Xu, Kate Saenko, and Bryan~A Plummer.
\newblock Logan: Latent graph co-attention network for weakly-supervised video moment retrieval.
\newblock In \emph{Proceedings of the IEEE/CVF Winter Conference on Applications of Computer Vision}, pages 2083--2092, 2021.

\bibitem[Vaswani et~al.(2017)Vaswani, Shazeer, Parmar, Uszkoreit, Jones, Gomez, Kaiser, and Polosukhin]{vaswani2017transformer}
Ashish Vaswani, Noam Shazeer, Niki Parmar, Jakob Uszkoreit, Llion Jones, Aidan~N Gomez, Lukasz Kaiser, and Illia Polosukhin.
\newblock Attention is all you need.
\newblock \emph{arXiv preprint arXiv:1706.03762}, 2017.

\bibitem[Wang et~al.(2021)Wang, Wang, Wu, Li, and Wu]{wang2021mmn}
Zhenzhi Wang, Limin Wang, Tao Wu, Tianhao Li, and Gangshan Wu.
\newblock Negative sample matters: A renaissance of metric learning for temporal grounding.
\newblock \emph{arXiv preprint arXiv:2109.04872}, 2021.

\bibitem[Wu et~al.(2020)Wu, Li, Han, and Lin]{wu2020bar}
Jie Wu, Guanbin Li, Xiaoguang Han, and Liang Lin.
\newblock Reinforcement learning for weakly supervised temporal grounding of natural language in untrimmed videos.
\newblock In \emph{Proceedings of the 28th ACM International Conference on Multimedia}, pages 1283--1291, 2020.

\bibitem[Xiao et~al.(2021)Xiao, Chen, Shao, Zhuang, and Xiao]{xiao2021lpnet}
Shaoning Xiao, Long Chen, Jian Shao, Yueting Zhuang, and Jun Xiao.
\newblock Natural language video localization with learnable moment proposals.
\newblock \emph{arXiv preprint arXiv:2109.10678}, 2021.

\bibitem[Xu et~al.(2019)Xu, Chen, Cheng, Duan, and Luo]{xu2019DAVQA}
Yiming Xu, Lin Chen, Zhongwei Cheng, Lixin Duan, and Jiebo Luo.
\newblock Open-ended visual question answering by multi-modal domain adaptation.
\newblock \emph{arXiv preprint arXiv:1911.04058}, 2019.

\bibitem[Yu et~al.(2018)Yu, Lin, Shen, Yang, Lu, Bansal, and Berg]{yu2018mattnet}
Licheng Yu, Zhe Lin, Xiaohui Shen, Jimei Yang, Xin Lu, Mohit Bansal, and Tamara~L. Berg.
\newblock Mattnet: Modular attention network for referring expression comprehension.
\newblock In \emph{Proceedings of the IEEE Conference on Computer Vision and Pattern Recognition}, June 2018.

\bibitem[Yuan et~al.(2019)Yuan, Mei, and Zhu]{yuan2019find}
Yitian Yuan, Tao Mei, and Wenwu Zhu.
\newblock To find where you talk: Temporal sentence localization in video with attention based location regression.
\newblock In \emph{Proceedings of the AAAI Conference on Artificial Intelligence}, volume~33, pages 9159--9166, 2019.

\bibitem[Yuan et~al.(2021)Yuan, Lan, Chen, Liu, Wang, and Zhu]{yuan2021ood}
Yitian Yuan, Xiaohan Lan, Long Chen, Wei Liu, Xin Wang, and Wenwu Zhu.
\newblock A closer look at temporal sentence grounding in videos: Datasets and metrics.
\newblock \emph{arXiv preprint arXiv:2101.09028}, 2021.

\bibitem[Zeng et~al.(2021)Zeng, Cao, Wei, Liu, Zhao, and Qin]{zeng2021mmrg}
Yawen Zeng, Da~Cao, Xiaochi Wei, Meng Liu, Zhou Zhao, and Zheng Qin.
\newblock Multi-modal relational graph for cross-modal video moment retrieval.
\newblock In \emph{Proceedings of the IEEE/CVF Conference on Computer Vision and Pattern Recognition}, pages 2215--2224, 2021.

\bibitem[Zhang et~al.(2019)Zhang, Dai, Wang, Wang, and Davis]{zhang2019man}
Da~Zhang, Xiyang Dai, Xin Wang, Yuan-Fang Wang, and Larry~S Davis.
\newblock Man: Moment alignment network for natural language moment retrieval via iterative graph adjustment.
\newblock In \emph{Proceedings of the IEEE/CVF Conference on Computer Vision and Pattern Recognition}, pages 1247--1257, 2019.

\bibitem[Zhang et~al.(2020{\natexlab{a}})Zhang, Peng, Fu, and Luo]{zhang20202dtan}
Songyang Zhang, Houwen Peng, Jianlong Fu, and Jiebo Luo.
\newblock Learning 2d temporal adjacent networks for moment localization with natural language.
\newblock In \emph{Proceedings of the AAAI Conference on Artificial Intelligence}, volume~34, pages 12870--12877, 2020{\natexlab{a}}.

\bibitem[Zhang et~al.(2020{\natexlab{b}})Zhang, Lin, Zhao, Zhu, and He]{zhang2020rtbpn}
Zhu Zhang, Zhijie Lin, Zhou Zhao, Jieming Zhu, and Xiuqiang He.
\newblock Regularized two-branch proposal networks for weakly-supervised moment retrieval in videos.
\newblock In \emph{Proceedings of the 28th ACM International Conference on Multimedia}, pages 4098--4106, 2020{\natexlab{b}}.

\bibitem[Zhang et~al.(2020{\natexlab{c}})Zhang, Zhao, Lin, He, et~al.]{zhang2020ccl}
Zhu Zhang, Zhou Zhao, Zhijie Lin, Xiuqiang He, et~al.
\newblock Counterfactual contrastive learning for weakly-supervised vision-language grounding.
\newblock \emph{Advances in Neural Information Processing Systems}, 33, 2020{\natexlab{c}}.

\end{thebibliography}
\end{document}